\documentclass[letterpaper]{article}
\usepackage{aaai2027}
\usepackage[hyphens]{url}
\usepackage{graphicx}
\usepackage{natbib}
\usepackage{caption}
\usepackage{algorithm}
\usepackage{algorithmic}
\usepackage{amsmath}
\usepackage{amssymb}
\usepackage{amsthm}
\usepackage{booktabs}
\usepackage{multirow}
\usepackage{bm}
\newtheorem{proposition}{Proposition}
\newtheorem{theorem}{Theorem}
\newtheorem{corollary}{Corollary}
\title{When Dynamics Models Read the Wrong Time Steps: \\ Label-Free Event Credit Re-Anchoring for Robust Global Readouts}
\author{Yifan Wang}
\affiliations{Department of Mechanical Engineering, McGill University\\ Montr\'eal, QC H3A 2T7, Canada\\ yifan.wang18@mail.mcgill.ca}

\begin{document}
\maketitle

\begin{abstract}
Learned dynamics models often answer global physical questions, such as fault severity or impact stiffness, by pooling a per-step feature sequence into one readout vector. This sequence-to-global interface creates an under-studied temporal credit problem: with only trajectory-level supervision, a model can predict accurately in training conditions while reading from abundant smooth correlates rather than the brief physical events that determine the target. We call this failure temporal credit dilution. It is not exposed by the training loss and is not removed by standard physics-informed residuals, because the error lies in where the global readout assigns functional credit. We introduce Credit-in-Event, an interface-level probe for measuring how much pooled credit lands on event steps, and prove in closed form that a pooled linear reader routes credit to a spurious background channel as the event fraction shrinks. We then propose CREST, a training-free and label-free readout that estimates a transient event core from learned features and re-anchors the pooled representation through event-versus-rest contrast. Across simulated gear and impact systems, recurrent and attention encoders, and public bearing vibration data, CREST reduces out-of-distribution error while restoring event credit. Ablations show that stable-step selection and receptive-field shrinking fail, confirming that the gain comes from event-core credit re-anchoring rather than a generic locality or stability prior.
\end{abstract}

\section{Introduction}

Learned models of physical dynamics increasingly answer global questions from a sequence of measurements: the severity of a developing fault, the stiffness revealed by an impact, a stability margin, or a safety score \citep{raissi2019physics,chen2018neural}. These deployments share a structure. A per-step encoder maps the trajectory to a feature sequence, an aggregation step pools that sequence into one vector, and a linear head reads off the scalar of interest \citep{vaswani2017attention,cho2014learning}. They also share a requirement, namely robustness when the operating condition changes between training and deployment.

The pooling step is where this structure becomes fragile. Supervision is coarse, with one label per trajectory, so the loss never states which time steps the answer should come from. We say a readout assigns \emph{functional credit} to a step when that step materially shapes the pooled vector and hence the prediction. When the decisive physical content is concentrated in a few brief events, while a smooth global statistic happens to correlate with the label during data collection, a pooled readout can lower its training error by placing its functional credit on the smooth statistic. We call this failure temporal credit dilution: the events that physically determine the target receive little credit in the pooled representation, and the prediction rests on a correlate that is not causal.

\begin{figure*}[t]
\centering
\includegraphics[width=\textwidth]{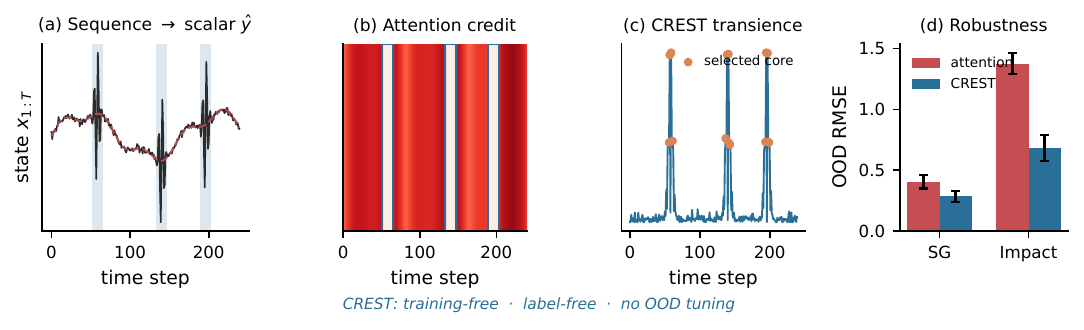}
\caption{Temporal credit dilution and CREST. (a) A sequence of physical states is compressed to one scalar $\hat{y}$; the target is set by sparse events (shaded), yet a smooth cue is globally predictive in training. (b) The attention readout places most credit on the smooth background and little on the event windows. (c) CREST scores feature transience by a low-pass residual, which peaks on the events, and selects a sparse core. (d) Under a shift that breaks the cue, CREST lowers out-of-distribution error on both systems, using no event labels.}
\label{fig:teaser}
\end{figure*}

This failure is invisible from the usual signals. The in-distribution error decreases smoothly, so validation curves look healthy. A physics-informed residual constrains the local state evolution but leaves the global readout free to summarize the wrong part of the trajectory, so it does not remove the failure, and the remedy we propose is not a new physics loss. Only under a distribution shift that breaks the smooth correlate does the error rise, and by then the model has committed to reading the wrong steps. Figure~\ref{fig:teaser} illustrates the phenomenon and our remedy.

Existing tools do not target this interface. Shortcut and spurious-correlation methods either act on input features or require group or environment labels, which are rarely available for a single stream of physical measurements \citep{geirhos2020shortcut,arjovsky2019invariant,sagawa2020distributionally,kirichenko2023last}. Shortcut studies in time series so far detect input-level point shortcuts through gradients \citep{ibarra2025gradient}, an interface that, as we prove, cannot see credit misallocated during pooling. Aggregation artifacts have also been observed in visual representations \citep{shi2026vision}, but global physical readouts pose a different question: whether a scalar prediction receives functional credit from the transient events that determine the measured quantity. This distinction motivates a temporal credit analysis rather than a visual-token repair.

We address temporal credit dilution directly at the readout. We introduce Credit-in-Event, an interface-level probe that quantifies how much pooled credit lands on the events, prove why pooling dilutes that credit, and propose Credit RE-anchoring through Sparse Transient readout (CREST), a training-free and label-free rule that re-anchors the pooled vector onto a transient event core estimated from the features themselves. Our contributions are as follows.

\begin{itemize}
\item \textbf{Temporal readout failure.} We identify temporal credit dilution, where a sequence-to-global dynamics model predicts accurately in distribution while assigning little readout credit to the sparse physical events that determine the target.
\item \textbf{Interface-level probe and theory.} We introduce Credit-in-Event and prove, in a sparse-event two-channel model, that global pooling makes event credit scale as $\Theta(\varepsilon^2)$ while an abundant background cue remains high signal-to-noise, so input-gradient saliency cannot detect the failure.
\item \textbf{Label-free event re-anchoring.} We propose CREST, a training-free readout that estimates a transient event core from learned features and forms an event-versus-rest contrast, without event labels, out-of-distribution labels, or per-system test tuning.
\item \textbf{Evidence and falsification.} Across two simulators, recurrent and attention encoders, and a public bearing dataset, CREST improves out-of-distribution error and restores event credit. Stable-step selection, receptive-field shrinking, and group robustness fail, isolating the temporal mechanism.
\end{itemize}

\section{Related Work}

\paragraph{Shortcut learning and out-of-distribution robustness.}
Deep models exploit cues that are predictive in training but not causal \citep{geirhos2020shortcut,nam2020learning}. Standard remedies assume group or environment structure, as in invariant risk minimization \citep{arjovsky2019invariant} and group distributionally robust optimization \citep{sagawa2020distributionally}, or retrain a final layer on a reweighted split \citep{kirichenko2023last}. Such structure is rarely available for a single stream of physical measurements, where there are no demographic groups and the operating condition that drives the shift is itself unobserved at deployment. We therefore intervene at the readout under a single training distribution.

\paragraph{Attribution and credit in time series.}
Saliency explains predictions through input sensitivity \citep{sundararajan2017axiomatic}, and recent work detects point-level shortcuts in time-series classification using gradients \citep{ibarra2025gradient}. Our Corollary~\ref{cor:interface} shows that the relevant failure is invisible to input sensitivity, because a step can be input-sensitive while contributing almost nothing to the pooled vector. Credit-in-Event is therefore not an attention-weight explanation but an interface measurement, and whether attention weights explain predictions is itself contested \citep{jain2019attention,wiegreffe2019attention}; we validate credit causally by masking.

\paragraph{Physics-informed dynamics learning.}
Physics-informed objectives and neural differential equations improve learned dynamics by constraining local evolution or residual consistency \citep{raissi2019physics,chen2018neural}. These constraints are valuable but orthogonal to the question studied here: after a feature sequence has been learned, a global readout can still place its functional credit on time steps that are predictive rather than physically decisive. CREST therefore targets the aggregation interface rather than the state-transition model or the residual loss.

\paragraph{Aggregation artifacts beyond dynamics.}
Vision Transformers can develop high-norm background tokens that serve internal aggregation roles, and recent work connects part of this behavior to shortcut-like aggregation under coarse semantic supervision \citep{darcet2024vision,shi2026vision}. CREST studies a different interface: a sequence of physical states compressed into one scalar readout, where the question is whether the pooled representation assigns credit to sparse events that determine the target. The decisive content is a short physical transient rather than a foreground region, the useful selection direction is transient rather than stable, and the analysis gives a closed-form sparse-event credit theorem and a budget law for event anchoring. We therefore treat visual-token artifact methods as related motivation, not as the mechanism or baseline solution for learned dynamics.

\paragraph{Selective pooling and weak supervision.}
Selecting informative elements appears in multiple-instance learning \citep{ilse2018attention}, typically for weakly supervised classification. CREST selects spectrally transient steps for out-of-distribution physical regression, ties the selection budget to a closed-form budget law, and runs without labels or training.

\section{Problem Setup and Credit Probe}

We observe trajectories $X=(x_1,\dots,x_T)$ with a coarse label $y\in\mathbb{R}$, one per trajectory. A physical event set $E\subset\{1,\dots,T\}$ with $L=|E|$ and event fraction $\varepsilon=L/T$ marks the steps that determine $y$. For simulators $E$ is known from the generating equations; for real vibration data $E$ is a diagnostic event window estimated from order tracking and used only for analysis and supervised references, and CREST never observes it. The model family is a per-step encoder $F_t=\phi(x_t)\in\mathbb{R}^{D}$, an aggregation $p=\mathrm{Agg}(F_{1:T})$, and a linear readout $\hat{y}=w^\top p$. The experimental variable is $\mathrm{Agg}$.

We measure credit at the aggregation interface. The model-agnostic step score is the cosine alignment between a step feature and the pooled vector,
\begin{equation}
s_t=\frac{\langle F_t,\,p\rangle}{\lVert F_t\rVert\,\lVert p\rVert},\qquad c_t=\max(s_t,0).
\end{equation}
The positive part $c_t$ discards steps whose feature direction opposes the pooled direction, since such steps do not support the final readout vector. We define
\begin{equation}
\mathrm{ECM}=\frac{\sum_{t\in E}c_t}{\sum_{t=1}^{T}c_t+\delta},\quad
\mathrm{Prec}@|E|=\frac{|\mathrm{Top}_{|E|}(s)\cap E|}{|E|},
\end{equation}
with $\mathrm{CiE}@1=\Pr(\arg\max_t s_t\in E)$ and a small $\delta>0$. The chance level for $\mathrm{CiE}@1$ equals $\varepsilon$ and is shown with every figure. All Credit-in-Event quantities are used only for diagnosis and reporting; they are not training objectives, and CREST does not use them for selection.

\section{Temporal Credit Dilution}

\paragraph{A solvable model.}
We use a two-channel generative model with fixed horizon $T$, event set $E$ of size $L=\varepsilon T$ independent of the noise, label $y\sim\mathcal{N}(0,1)$, independent unit Gaussians $\xi$, noise scales $s_0,s_1>0$, and cue strength $\gamma\in(0,1]$:
\begin{align}
x^{0}_t &= \mathbf{1}[t\in E]\,y + s_0\,\xi^{0}_t, \\
x^{1}_t &= \mathbf{1}[t\notin E]\,(g\,y) + \mathbf{1}[t\notin E]\,s_1\,\xi^{1}_t .
\end{align}
Channel $0$ is the invariant event channel and channel $1$ the spurious background, with $g=\gamma$ in distribution and $g=0$ out of distribution while the noise law is preserved. Risks are normalized mean squared error with $\mathrm{Var}(y)=1$. A pooled linear reader sees only the global means $m_j=\frac{1}{T}\sum_t x^{j}_t$, and
\begin{equation}
S_E=\frac{\varepsilon^2 T}{s_0^2}, \quad S_B=\frac{(1-\varepsilon)\gamma^2 T}{s_1^2}, \quad S=S_E+S_B .
\end{equation}
The event channel is observed only on $\varepsilon T$ steps, and global averaging shrinks its amplitude by $\varepsilon$, so its pooled signal power scales as $\varepsilon^2$, whereas the background cue occupies $(1-\varepsilon)T$ steps and stays order one.

\begin{proposition}[Sparse-event credit dilution]
\label{prop:existence}
Consider $\varepsilon\to 0$ with $T$ fixed and $S_B$ bounded away from zero. The population least-squares reader over $(m_0,m_1)$ has
\begin{equation}
R_{\mathrm{id}}=\frac{1}{1+S},\qquad
R_{\mathrm{ood}}=\Big(\tfrac{1+S_B}{1+S}\Big)^{2}+\frac{S}{(1+S)^2},
\end{equation}
and event credit share $\rho_E=S_E/S=\Theta(\varepsilon^2)$. Hence
\begin{equation}
\lim_{\varepsilon\to 0} R_{\mathrm{ood}} = 1+\frac{S_B}{(1+S_B)^2} > 1,\quad
\lim_{\varepsilon\to 0} R_{\mathrm{id}} = \frac{1}{1+S_B}.
\end{equation}
\end{proposition}

A high signal-to-noise background cue therefore makes the in-distribution risk small while the same pooled reader becomes worse than predicting the mean once the cue is removed. Proposition~\ref{prop:existence} establishes that the failure can be the population-optimal pooled solution in the sparse-event regime: the model appears reliable in distribution, while the event credit vanishes as $\Theta(\varepsilon^2)$ and the out-of-distribution risk exceeds the mean predictor when the cue breaks. Proofs are in the supplement, where the predicted risks match an empirical fit to three decimals.

\begin{corollary}[Interface-measurement principle]
\label{cor:interface}
The per-step input sensitivity of the same reader has ratio $w_0^\star/w_1^\star=\varepsilon\,s_1^2/(\gamma\,s_0^2)$ on event versus background steps, which is order one and can exceed one. Input-gradient saliency can therefore identify event steps as locally sensitive while failing to reveal that their aggregate contribution to the pooled readout is negligible.
\end{corollary}

\begin{figure}[t]
\centering
\includegraphics[width=0.88\columnwidth]{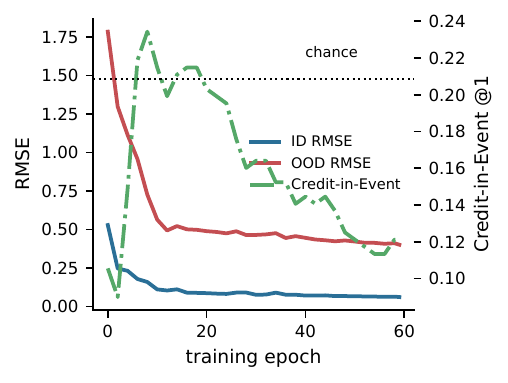}
\caption{Training can lower loss while event credit disappears. On the gear system, in-distribution error decreases, but Credit-in-Event peaks early and then falls below the event-rate chance line while out-of-distribution error remains high.}
\label{fig:dynamics}
\end{figure}

\paragraph{Training-dynamics fingerprint.}
Figure~\ref{fig:dynamics} tracks an attention readout during training on the gear system. The in-distribution error falls by nearly an order of magnitude while the out-of-distribution error stalls, and the event credit rises briefly and then decays below the chance line, as the model re-routes credit onto the cheaper global average.

\begin{figure}[t]
\centering
\includegraphics[width=\columnwidth]{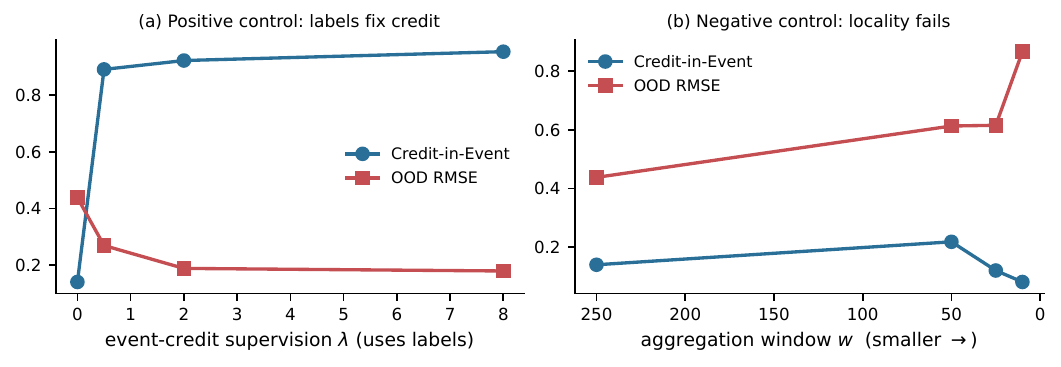}
\caption{Two-factor isolation. (a) Positive control: event-localized supervision raises credit and lowers error but requires labels. (b) Negative control: shrinking the aggregation window does not move credit onto the events, so a locality prior is insufficient.}
\label{fig:isolation}
\end{figure}

\paragraph{Two factors, isolated.}
Figure~\ref{fig:isolation} separates the causes. Adding an event-localized auxiliary loss lifts event credit and halves the out-of-distribution error, a positive control that needs event labels. Shrinking the aggregation window does not move credit onto the events and hurts both errors, a negative control showing that a locality prior cannot starve a cue that is locally readable. A selection-based remedy is therefore needed.

\section{CREST}

CREST re-anchors the pooled representation onto the transient steps without labels or training. For channel $j$ we compute a normalized low-pass residual and an averaged transience profile,
\begin{equation}
r_{tj}=\frac{|F_{tj}-\mathrm{LP}_\sigma(F_{\cdot j})_t|}{\sum_u|F_{uj}-\mathrm{LP}_\sigma(F_{\cdot j})_u|+\delta},\qquad
r_t=\frac{1}{D}\sum_{j=1}^{D}r_{tj},
\end{equation}
where $\mathrm{LP}_\sigma$ is a Gaussian low-pass along time in the real Fourier domain. A high $r_{tj}$ marks transient, event-like content.

\paragraph{Label-free budget and contrast.}
We normalize the averaged profile to $S_t\in[0,1]$ per trajectory, set $P_t=S_t/(\sum_u S_u+\delta)$, and read its concentration through a participation width $b=\frac{1}{T}\big(\sum_t P_t^2\big)^{-1}$, where small $b$ means a sharp profile. The budget blends a sharp-core estimate with an Otsu-thresholded tail estimate \citep{otsu1979threshold},
\begin{align}
\varepsilon_{\mathrm{core}}&=\mathrm{clip}(0.55\,b-0.17,\,\varepsilon_{\min},\,\varepsilon_{\max}),\\
g_b&=\mathrm{clip}\!\Big(\tfrac{b-0.45}{0.10},0,1\Big),\quad
\hat\varepsilon=(1-g_b)\,\varepsilon_{\mathrm{core}}+g_b\,\varepsilon_{\mathrm{tail}},
\end{align}
and the contrast weight follows the same width,
\begin{equation}
\alpha=\mathrm{clip}(1.025-2.625\,\hat\varepsilon,\,0.50,\,1.00).
\end{equation}
Sharp cores receive a stronger event-versus-rest contrast and broad events more global context. All constants are fixed before evaluation and shared across systems; exact values are in the supplement.

\paragraph{Selective readout.}
For channel $j$ we select the top $K'=\lceil\hat\varepsilon T/(2d+1)\rceil$ entries of $r_{tj}$ and dilate each by radius $d$ to form a mask $M_{tj}$ covering about $\hat\varepsilon T$ steps, with $d=2$. The masked means $\bar F_{\mathrm{sel}},\bar F_{\mathrm{rest}},\bar F_{\mathrm{global}}$ over the selected, remaining, and all steps give
\begin{equation}
p=\alpha\big(\bar F_{\mathrm{sel}}-\bar F_{\mathrm{rest}}\big)+(1-\alpha)\,\bar F_{\mathrm{global}}.
\label{eq:crest}
\end{equation}
The selection is non-differentiable and uses a stop-gradient, so gradients flow only through the gathered features \citep{bengio2013estimating}; the per-step vote count gives a free, training-free event localization and is never a source of labels. Because the selector reads the learned features it is not strictly noise-independent; a sample-splitting variant in the supplement restores independence.

\begin{algorithm}[t]
\caption{CREST readout: label-free transient-core re-anchoring}
\label{alg:crest}
\begin{algorithmic}[1]
\REQUIRE step features $F\in\mathbb{R}^{T\times D}$, low-pass width $\sigma$, dilation radius $d$
\STATE compute channel residuals $r_{tj}$ and averaged transience $r_t$
\STATE estimate $\hat\varepsilon$ from the concentration of $r_t$; set $\alpha=\mathrm{clip}(1.025-2.625\hat\varepsilon,0.50,1.00)$
\STATE $K'\leftarrow\lceil \hat\varepsilon T/(2d+1)\rceil$
\STATE for each channel $j$, select the top-$K'$ indices of $r_{tj}$ and dilate by radius $d$ to obtain $M_{tj}$
\STATE compute $\bar F_{\mathrm{sel}},\bar F_{\mathrm{rest}},\bar F_{\mathrm{global}}$ using $M$
\STATE $p\leftarrow\alpha(\bar F_{\mathrm{sel}}-\bar F_{\mathrm{rest}})+(1-\alpha)\bar F_{\mathrm{global}}$
\RETURN pooled vector $p$; stop gradients through $M$
\end{algorithmic}
\end{algorithm}

\paragraph{Recovery and budget law.}
Let an idealized selector return $\hat E$ of effective support $K$ and precision $\pi=|\hat E\cap E|/K$, independent of the noise, and let $L$ be the true event-core support.

\begin{proposition}[Anchored recovery]
\label{prop:recovery}
The anchored event signal-to-noise ratio at full anchoring is $\pi^2K/s_0^2$, linear in $K$ rather than quadratic in $\varepsilon$, and the out-of-distribution risk decreases in this quantity, which increases in the anchoring weight whenever $\pi>\varepsilon$.
\end{proposition}

\begin{theorem}[Budget law]
\label{thm:budget}
With the best selector of size $K$ the risk is $R(K)=s_0^2/(\pi(K)^2K+s_0^2)$, U-shaped with its minimum at $K=L$. At $K=T$ the anchored event signal-to-noise ratio reduces to $\varepsilon^2T$, the same dilution as the event channel of global averaging.
\end{theorem}

Theorem~\ref{thm:budget} identifies the target budget for an ideal selector; the label-free estimator is an approximation, and Figure~\ref{fig:ksweep} tests whether its selected budget lies in the low-risk regime.

\section{Experiments}

\paragraph{Systems and splits.}
SG-Drive is a lumped torsional gear-drive simulator with time-varying mesh stiffness and sparse faulty-tooth engagement events; the target is a single-tooth fault severity, with a collection bias in which damaged units are operated gently so a smooth speed level becomes spuriously predictive, and the out-of-distribution split removes the bias. The impact oscillator is a wall-contact system whose target is the log wall stiffness, with the out-of-distribution split decorrelating the drive amplitude from stiffness. For each simulator we use $1024$ training, $256$ in-distribution, and $256$ out-of-distribution trajectories, and a hand-designed event-window estimator verifies that the target is decodable from the event windows. CWRU provides public drive-end inner-race bearing vibration \citep{smith2015rolling}; we train on motor loads $0,1,2$ and test on the held-out load $3$, with event windows from order tracking used only for probes and the supervised reference.

\paragraph{Baselines and metrics.}
We compare CREST against mean pooling, attention pooling, register-augmented attention, windowed attention, and a supervised event-window reference that pools the annotated event steps, together with last-state pooling, empirical risk minimization, and group distributionally robust optimization on a recurrent encoder. The group-robust baseline uses quartiles of the known smooth speed proxy as groups, giving it access to the spurious factor but not to event windows. We report in-distribution and out-of-distribution root mean squared error, Credit-in-Event, the precision at the event rate, and the event credit mass, never reporting an out-of-distribution number without its in-distribution counterpart. We use ten seeds for the headline systems and five elsewhere, with one-sided paired signed-rank tests against attention. All CREST constants are fixed before out-of-distribution evaluation and shared across systems; out-of-distribution labels are never used for method selection.

\begin{figure}[t]
\centering
\includegraphics[width=\columnwidth]{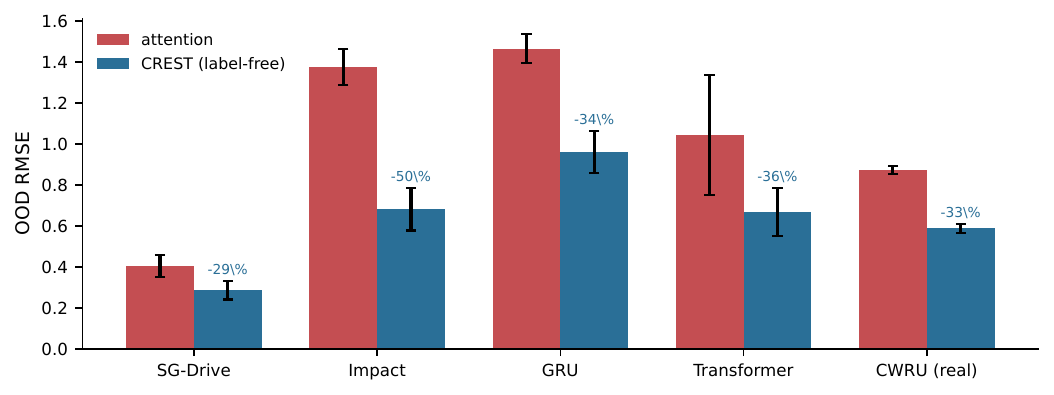}
\caption{Cross-system out-of-distribution error of attention and label-free CREST across two simulators, two encoder families, and a public held-load bearing split, with relative improvements annotated and one standard deviation over seeds. In-distribution counterparts are in Table~\ref{tab:main} and the supplement.}
\label{fig:cross}
\end{figure}

\paragraph{Main result.}
Table~\ref{tab:main} reports the headline systems and Figure~\ref{fig:cross} the cross-system summary. A single label-free CREST configuration improves out-of-distribution error by $29\%$ on SG-Drive and $50\%$ on the impact system relative to attention, with paired signed-rank significance over ten seeds ($p<0.001$ on both), and it roughly doubles to triples the event credit. The improvement also appears on the public CWRU bearing benchmark under a held-load split, where CREST reduces held-load error from $0.874$ to $0.589$ over five seeds. We treat this as public real-data evidence rather than a complete field validation, since it covers one bearing benchmark and one held-load protocol.

\begin{table}[t]
\centering
\small
\setlength{\tabcolsep}{3pt}
\begin{tabular}{llccccc}
\toprule
System & Method & ID & OOD & $\Delta$OOD & CiE@1 & ECM \\
\midrule
\multirow{2}{*}{SG-Drive} & attention & $0.052$ & $0.404$ & -- & $0.27$ & $0.26$ \\
 & \textbf{CREST} & {\footnotesize $0.072$} & $\mathbf{0.286}$ & $-29\%$ & $\mathbf{0.62}$ & $\mathbf{0.62}$ \\
\midrule
\multirow{2}{*}{Impact} & attention & $0.173$ & $1.377$ & -- & $0.02$ & $0.04$ \\
 & \textbf{CREST} & {\footnotesize $0.160$} & $\mathbf{0.683}$ & $-50\%$ & $\mathbf{0.80}$ & $\mathbf{0.86}$ \\
\bottomrule
\end{tabular}
\caption{Headline results over ten seeds (means; standard deviations: SG-Drive OOD $0.404{\pm}0.055$ versus $0.286{\pm}0.045$; Impact OOD $1.377{\pm}0.087$ versus $0.683{\pm}0.105$). CREST uses no event labels, out-of-distribution labels, or per-system test tuning; paired one-sided signed-rank tests give $p<0.001$ for both. Full statistics are in the supplement.}
\label{tab:main}
\end{table}

\paragraph{Stronger baselines.}
On the gear system with a recurrent encoder, last-state pooling, empirical risk minimization, and group distributionally robust optimization all reach an out-of-distribution error near $1.46$ with event credit at the mean-pool floor, while CREST reaches $0.96$. Group robustness over the smooth proxy does not repair temporal credit.

\begin{figure}[t]
\centering
\includegraphics[width=\columnwidth]{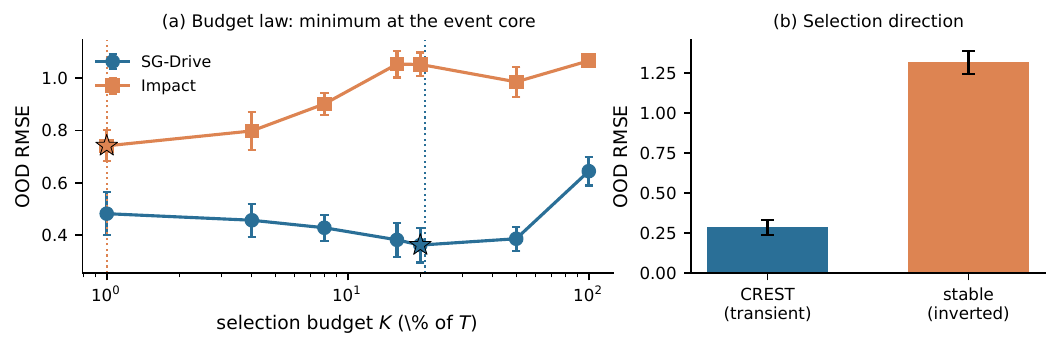}
\caption{Ablations. (a) Out-of-distribution error is U-shaped in the selection budget, with the minimum tracking the event core rather than a universal fraction, which supports the adaptive estimator. (b) Selecting stable steps inverts the direction and is the worst configuration.}
\label{fig:ksweep}
\end{figure}

\paragraph{Budget law and direction.}
Figure~\ref{fig:ksweep} traces the budget law and the selection direction. The out-of-distribution error is U-shaped in the budget, with the minimum near the dilated event rate on the gear system and at a narrower core on the impact system, consistent with Theorem~\ref{thm:budget} once the annotation width is accounted for. Selecting stable steps rather than transient ones is the worst configuration in the study, which is the key evidence that CREST is not a stability-based visual-token repair applied to time.

\begin{figure}[t]
\centering
\includegraphics[width=0.78\columnwidth]{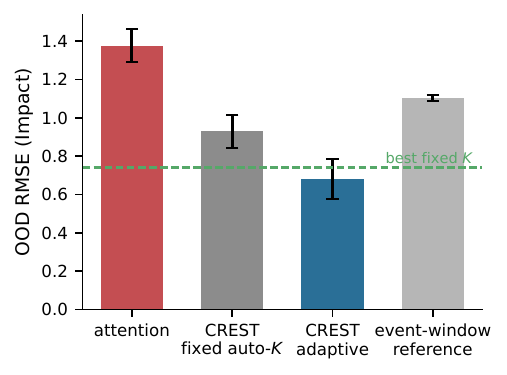}
\caption{Label-free core adaptation on the impact system. The fixed automatic budget includes halo steps, while the adaptive core estimator selects a narrower contact core and lowers error. The dashed line is the best fixed budget on the evaluated grid, and the reference pools the dilated diagnostic annotation rather than an ideal core.}
\label{fig:adaptive}
\end{figure}

\paragraph{Label-free adaptation.}
Figure~\ref{fig:adaptive} isolates the estimator on the impact system. A fixed automatic budget reaches $0.93$, while estimating the narrow contact core lowers the error to $0.68$, below the best fixed budget on the evaluated grid, and the same estimator leaves the gear result unchanged. On this system CREST also improves over the dilated event-window reference. This does not mean label supervision is inferior in principle; it indicates that the annotated window contains non-informative halo steps and that the informative core is narrower than the diagnostic window.

\paragraph{The restored credit is causal.}
We interpolate out the steps each readout rates as most important and re-evaluate. Removing the attention readout's highest-credit steps barely changes its error, and removing the true events is almost free for it, which confirms that it never used them. Removing the steps that CREST selects is catastrophic by comparison, so the two readouts rest on different parts of the trajectory. Curves are in the supplement, and Table~\ref{tab:claims} maps each claim to its evidence.

\begin{table}[t]
\centering
\small
\setlength{\tabcolsep}{3.5pt}
\begin{tabular}{p{4.9cm}p{2.6cm}}
\toprule
Claim & Evidence \\
\midrule
Pooling dilutes sparse-event credit & Prop.~\ref{prop:existence}, Supp.~B \\
Input saliency misses readout credit & Cor.~\ref{cor:interface} \\
Supervision fixes credit but needs labels & Fig.~\ref{fig:isolation}a \\
Receptive-field shrinking fails & Fig.~\ref{fig:isolation}b \\
CREST improves OOD without labels & Tab.~\ref{tab:main}, Fig.~\ref{fig:cross} \\
Adaptive core estimation matters & Fig.~\ref{fig:adaptive} \\
CREST reads causal steps & Fig.~S1, Supp.~I \\
\bottomrule
\end{tabular}
\caption{Claim-to-evidence map.}
\label{tab:claims}
\end{table}

\paragraph{Limitations.}
CREST is not a universal shortcut remedy. It is designed for sequence-to-global physical readouts where the decisive evidence is sparse and transient. If the target is determined by diffuse slow dynamics, or if the spurious cue is also transient and co-located with the event, the transience prior may not separate causal from non-causal evidence. The current real-data evidence covers one public bearing benchmark under one held-load protocol, and broader rotating-machinery and structural-impact studies are needed. Finally, the selector is computed from learned features, so the independence assumption in the recovery theorem is an idealization, and the supplement reports a sample-splitting variant that reduces this dependence.

\section{Conclusion}

We identified temporal credit dilution, a readout-level failure in which a learned dynamics model predicts accurately in distribution while reading from the wrong time steps, proved why global pooling dilutes the credit of sparse physical events, and introduced CREST, a training-free and label-free rule that estimates a transient event core and re-anchors the pooled representation. The remedy lowers out-of-distribution error and restores event credit across simulated and real contact systems and across encoder families. For global questions about physical dynamics, where the readout reads from deserves as much attention as the loss.

\bibliography{references}

@article{geirhos2020shortcut,
  title={Shortcut learning in deep neural networks},
  author={Geirhos, Robert and Jacobsen, J{\"o}rn-Henrik and Michaelis, Claudio and Zemel, Richard and Brendel, Wieland and Bethge, Matthias and Wichmann, Felix A},
  journal={Nature Machine Intelligence}, volume={2}, number={11}, pages={665--673}, year={2020},
  doi={10.1038/s42256-020-00257-z}
}

@inproceedings{darcet2024vision,
  title={Vision Transformers Need Registers},
  author={Darcet, Timoth{\'e}e and Oquab, Maxime and Mairal, Julien and Bojanowski, Piotr},
  booktitle={International Conference on Learning Representations (ICLR)}, year={2024}
}

@inproceedings{shi2026vision,
  title={Vision Transformers Need More Than Registers},
  author={Shi, Cheng and Yu, Yizhou and Yang, Sibei},
  booktitle={Proceedings of the IEEE/CVF Conference on Computer Vision and Pattern Recognition (CVPR)},
  pages={26328--26337},
  year={2026}
}

@inproceedings{ibarra2025gradient,
  title={Gradient-based Model Shortcut Detection for Time Series Classification},
  author={Ibarra, Salomon and Cantu, Frida and Zhou, Kaixiong and Zhang, Li},
  booktitle={2025 International Conference on Machine Learning and Applications (ICMLA)},
  pages={726--731},
  year={2025},
  doi={10.1109/ICMLA66185.2025.00104}
}

@article{arjovsky2019invariant,
  title={Invariant Risk Minimization},
  author={Arjovsky, Martin and Bottou, L{\'e}on and Gulrajani, Ishaan and Lopez-Paz, David},
  journal={arXiv preprint arXiv:1907.02893}, year={2019}
}

@inproceedings{sagawa2020distributionally,
  title={Distributionally Robust Neural Networks for Group Shifts: On the Importance of Regularization for Worst-Case Generalization},
  author={Sagawa, Shiori and Koh, Pang Wei and Hashimoto, Tatsunori B and Liang, Percy},
  booktitle={International Conference on Learning Representations (ICLR)}, year={2020}
}

@inproceedings{kirichenko2023last,
  title={Last Layer Re-Training is Sufficient for Robustness to Spurious Correlations},
  author={Kirichenko, Polina and Izmailov, Pavel and Wilson, Andrew Gordon},
  booktitle={International Conference on Learning Representations (ICLR)}, year={2023}
}

@inproceedings{nam2020learning,
  title={Learning from Failure: Training Debiased Classifier from Biased Classifier},
  author={Nam, Junhyun and Cha, Hyuntak and Ahn, Sungsoo and Lee, Jaeho and Shin, Jinwoo},
  booktitle={Advances in Neural Information Processing Systems (NeurIPS)}, year={2020}
}

@inproceedings{vaswani2017attention,
  title={Attention is All You Need},
  author={Vaswani, Ashish and Shazeer, Noam and Parmar, Niki and Uszkoreit, Jakob and Jones, Llion and Gomez, Aidan N and Kaiser, Lukasz and Polosukhin, Illia},
  booktitle={Advances in Neural Information Processing Systems (NeurIPS)}, year={2017}
}

@inproceedings{cho2014learning,
  title={Learning Phrase Representations using RNN Encoder-Decoder for Statistical Machine Translation},
  author={Cho, Kyunghyun and van Merri{\"e}nboer, Bart and Gulcehre, Caglar and Bahdanau, Dzmitry and Bougares, Fethi and Schwenk, Holger and Bengio, Yoshua},
  booktitle={Conference on Empirical Methods in Natural Language Processing (EMNLP)},
  pages={1724--1734},
  year={2014},
  doi={10.3115/v1/D14-1179}
}

@inproceedings{chen2018neural,
  title={Neural Ordinary Differential Equations},
  author={Chen, Ricky T. Q. and Rubanova, Yulia and Bettencourt, Jesse and Duvenaud, David},
  booktitle={Advances in Neural Information Processing Systems (NeurIPS)}, year={2018}
}

@article{raissi2019physics,
  title={Physics-informed neural networks: A deep learning framework for solving forward and inverse problems involving nonlinear partial differential equations},
  author={Raissi, Maziar and Perdikaris, Paris and Karniadakis, George E},
  journal={Journal of Computational Physics}, volume={378}, pages={686--707}, year={2019},
  doi={10.1016/j.jcp.2018.10.045}
}

@inproceedings{jain2019attention,
  title={Attention is not Explanation},
  author={Jain, Sarthak and Wallace, Byron C},
  booktitle={Conference of the North American Chapter of the Association for Computational Linguistics (NAACL)},
  pages={3543--3556},
  year={2019},
  doi={10.18653/v1/N19-1357}
}

@inproceedings{wiegreffe2019attention,
  title={Attention is not not Explanation},
  author={Wiegreffe, Sarah and Pinter, Yuval},
  booktitle={Conference on Empirical Methods in Natural Language Processing (EMNLP)},
  pages={11--20},
  year={2019},
  doi={10.18653/v1/D19-1002}
}

@article{smith2015rolling,
  title={Rolling element bearing diagnostics using the Case Western Reserve University data: A benchmark study},
  author={Smith, Wade A. and Randall, Robert B.},
  journal={Mechanical Systems and Signal Processing}, volume={64--65}, pages={100--131}, year={2015},
  doi={10.1016/j.ymssp.2015.04.021}
}

@inproceedings{ilse2018attention,
  title={Attention-based Deep Multiple Instance Learning},
  author={Ilse, Maximilian and Tomczak, Jakub M. and Welling, Max},
  booktitle={Proceedings of the 35th International Conference on Machine Learning (ICML)},
  series={PMLR}, volume={80}, pages={2127--2136}, year={2018}
}

@article{otsu1979threshold,
  title={A threshold selection method from gray-level histograms},
  author={Otsu, Nobuyuki},
  journal={IEEE Transactions on Systems, Man, and Cybernetics}, volume={9}, number={1}, pages={62--66}, year={1979},
  doi={10.1109/TSMC.1979.4310076}
}

@inproceedings{sundararajan2017axiomatic,
  title={Axiomatic Attribution for Deep Networks},
  author={Sundararajan, Mukund and Taly, Ankur and Yan, Qiqi},
  booktitle={International Conference on Machine Learning (ICML)}, year={2017}
}

@article{bengio2013estimating,
  title={Estimating or Propagating Gradients Through Stochastic Neurons for Conditional Computation},
  author={Bengio, Yoshua and L{\'e}onard, Nicholas and Courville, Aaron},
  journal={arXiv preprint arXiv:1308.3432}, year={2013}
}
\end{document}